%% file: main_acl.tex
\documentclass[11pt]{article}

\usepackage[preprint]{acl}

\usepackage{times}
\usepackage{latexsym}

\usepackage[T1]{fontenc}

\usepackage[utf8]{inputenc}

\usepackage{microtype}

\usepackage{inconsolata}

\usepackage{graphicx}

\usepackage[utf8]{inputenc} 
\usepackage[T1]{fontenc}    
\usepackage{hyperref}       
\usepackage{url}            
\usepackage{booktabs}       
\usepackage{amsfonts}       
\usepackage{nicefrac}       
\usepackage{microtype}      
\usepackage[dvipsnames]{xcolor}         
\usepackage{graphicx} 
\usepackage{parskip}  
\usepackage{setspace} 
\usepackage{refcount} 
\usepackage{upquote}  
\usepackage{fancyhdr}
\usepackage{lastpage}       
\usepackage{lscape}         
\usepackage{verbatim}       
\usepackage{listings}       
\usepackage{epsfig}         
\usepackage{array}          
\usepackage{enumitem}       
\usepackage{hhline}         
\usepackage{siunitx}        
\usepackage{amsmath}        
\usepackage{amssymb}        
\usepackage{amsthm}         
\usepackage{algorithm, algorithmicx, algpseudocode}
\usepackage{ifthen}         
\usepackage{float}
\newfloat{algorithm}{t}{lop}
\usepackage{cleveref}
\usepackage{booktabs}
\usepackage{makecell}

\usepackage{pdfpages}
\usepackage{subcaption}
\usepackage{multirow}
\usepackage{multicol}
\usepackage{wrapfig}
\usepackage[most]{tcolorbox} 
\usepackage{lipsum}

\title{Team of Thoughts: Efficient Test-time Scaling of Agentic Systems through Orchestrated Tool Calling}

\author{
Jeffrey T. H. Wong${^{1,\dagger}}$\quad
Zixi Zhang$^{1,\dagger}$\quad
Junyi Liu$^{2}$\quad
Yiren Zhao$^{1}$\\
$^{1}$Imperial College London, $^{2}$Microsoft Research\\
\texttt{\{tsz.wong20,b.zhang25,a.zhao\}@imperial.ac.uk}\\
\texttt{junyi.liu@microsoft.com} \\
}

\usepackage[textsize=tiny]{todonotes}

\begin{document}
\maketitle

\def\thefootnote{$\dagger$}\footnotetext{These authors contributed equally to this work.}\def\thefootnote{\arabic{footnote}}

\renewcommand{\thefootnote}{\fnsymbol{footnote}}
\begin{abstract}
  Existing Multi-Agent Systems (MAS) typically rely on homogeneous model configurations, failing to exploit the diverse expertise inherent in different post-trained architectures. We propose Team-of-Thoughts, a heterogeneous MAS framework that treats diverse models as specialized tools within an orchestrator-driven paradigm. Team-of-Thoughts introduces two novel components: (1) Orchestrator Calibration, which identifies models with superior coordination and synthesis capabilities, and (2) Agent Self-Assessment, a protocol where tool agents profile their own domain-specific strengths to guide selection. At inference, the orchestrator dynamically activates the most compatible agents based on these profiles to maximize capability coverage. Across five mathematical reasoning and code generation benchmarks, Team-of-Thoughts consistently outperforms individual models and existing MAS baselines. Notably, on AIME24 and LiveCodeBench, Team-of-Thoughts achieves 96.00\% and 77.91\% accuracy, respectively, significantly improving over homogeneous role-play baselines (80.00\% and 65.93\%). 
  \footnote[1]{Code is available at \url{https://github.com/JeffreyWong20/Team-of-Thoughts}.}
\end{abstract}



\input{sections/introduction}

\input{sections/background}

\input{sections/method}

\input{sections/experiment}

\input{sections/related}

\input{sections/conclusion}

\bibliography{references}

\newpage
\appendix
\input{sections/appendix}

\end{document}

%% file: sections/introduction.tex
\section{Introduction}

\input{figures/fig_accuracy_vs_cost}

\begin{figure*}[t]
    \centering
    \includegraphics[width=\linewidth]{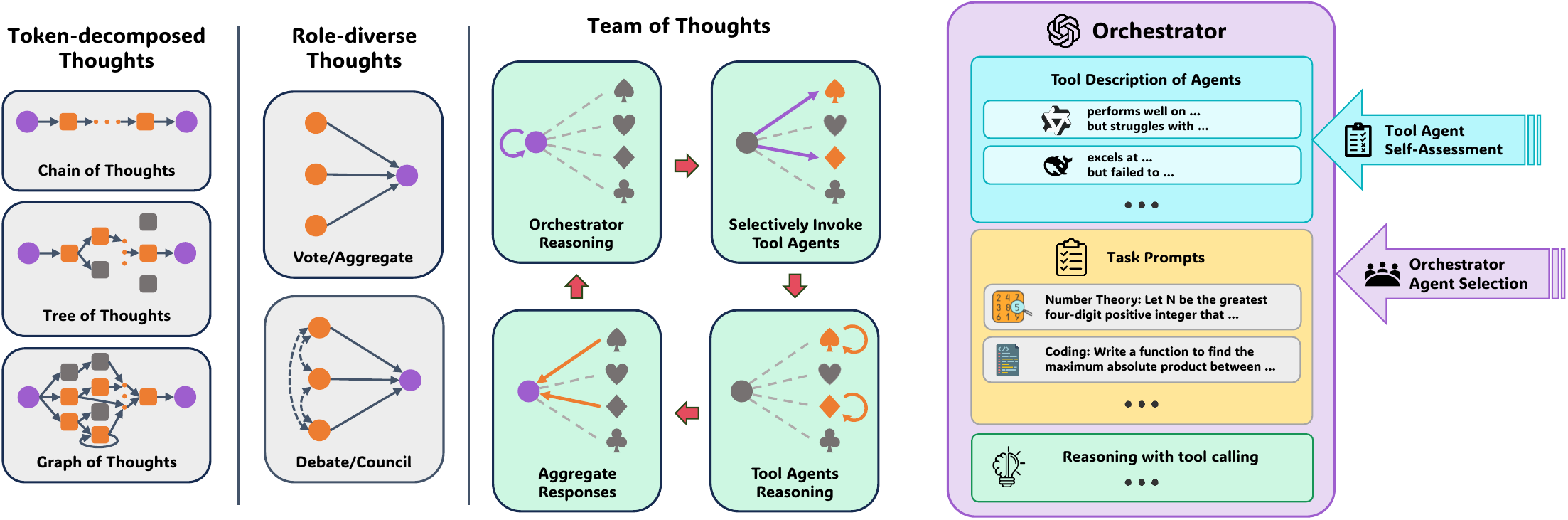}
    \caption{\textbf{Overview of Team-of-Thoughts.} \textbf{(Left)} Unlike single-model reasoning or homogeneous MAS (role-play), Team-of-Thoughts leverages heterogeneous model priors to maximize solution space coverage. \textbf{(Right)} Our architecture utilizes a pre-inference pipeline---comprising orchestrator calibration and agent self-profiling---to enable dynamic agent selection. During inference, the orchestrator selectively invokes optimal tool agents and synthesizes their outputs into a high-confidence final response.}
    \label{fig:overview}
\end{figure*}

Test-time scaling (TTS) has emerged as a critical paradigm for extending the capabilities of large language models (LLMs) beyond their training-time performance~\citep{ttsscaling, inferencescaling}. By allocating additional computational budget during inference via methods such as process reward model (PRM) scoring, beam search, or tree-based exploration~\citep{cot, tot, got}, models can unlock latent reasoning capabilities to solve complex tasks. This shift recognizes that the strategic deployment of inference-time compute is as fundamental to model performance as the scale of pre-training.

Despite these gains, current TTS approaches are typically confined to single-model execution or static multi-agent workflows with fixed role assignments for a single model family~\citep{chatdev, metagpt, autogen}. This homogeneity prevents systems from exploiting the complementary strengths inherent in diverse LLMs, which often possess divergent expertise due to distinct post-training procedures and dataset compositions. Recent industry developments, such as Grok 4.2~\citep{grok42} and Claude’s agent teams~\citep{anthropic_agent_teams_2026}, signal a nascent industry shift toward coordinated multi-agent reasoning. However, existing academic frameworks~\citep{CoA2024, agentverse} still lack the dynamic flexibility required to coordinate these diverse agents based on model-specific expertise and task attributes.

We address these limitations with \textbf{Team-of-Thoughts}, a novel Multi-Agent System (MAS) that achieves efficient test-time scaling through an orchestrated tool-calling paradigm as highlighted in \Cref{fig:motivation}. Rather than treating models as monolithic reasoners or assigning them rigid personas, our framework conceptualizes \textbf{diverse LLMs as specialized tools} that can be dynamically invoked. By leveraging the native tool-calling capabilities of modern LLMs, we construct a hierarchical architecture where a central orchestrator strategically activates the most suitable tool agents for a query.

This design facilitates massive parallelism and superior token efficiency. By replacing sequential, token-heavy reasoning with a coordinated team effort, Team-of-Thoughts ensures that the most proficient ``specialists'' are consulted at the optimal time. Our key contributions are:
\begin{itemize}\itemsep0em\parskip0.5em
    \item \textbf{Team-of-Thoughts Framework:} A novel MAS architecture enabling heterogeneous LLMs to collaborate through a dynamic, tool-calling hierarchy that maximizes capability coverage.
    \item \textbf{Orchestration and Self-Assessment Mechanisms:} We introduce an \textit{orchestration calibration} to identify superior coordinators and a \textit{self-assessment protocol} for agents to profile their own domain expertise.
    \item \textbf{Empirical Superiority:} Extensive evaluation showing Team-of-Thoughts consistently outperforms standalone models and MAS baselines. Notably, achieving \textbf{96.00\%} on AIME24 and \textbf{77.91\%} on LiveCodeBench.
\end{itemize}

%% file: figures/fig_accuracy_vs_cost.tex
\begin{figure}
    \centering
    \includegraphics[width=\columnwidth]{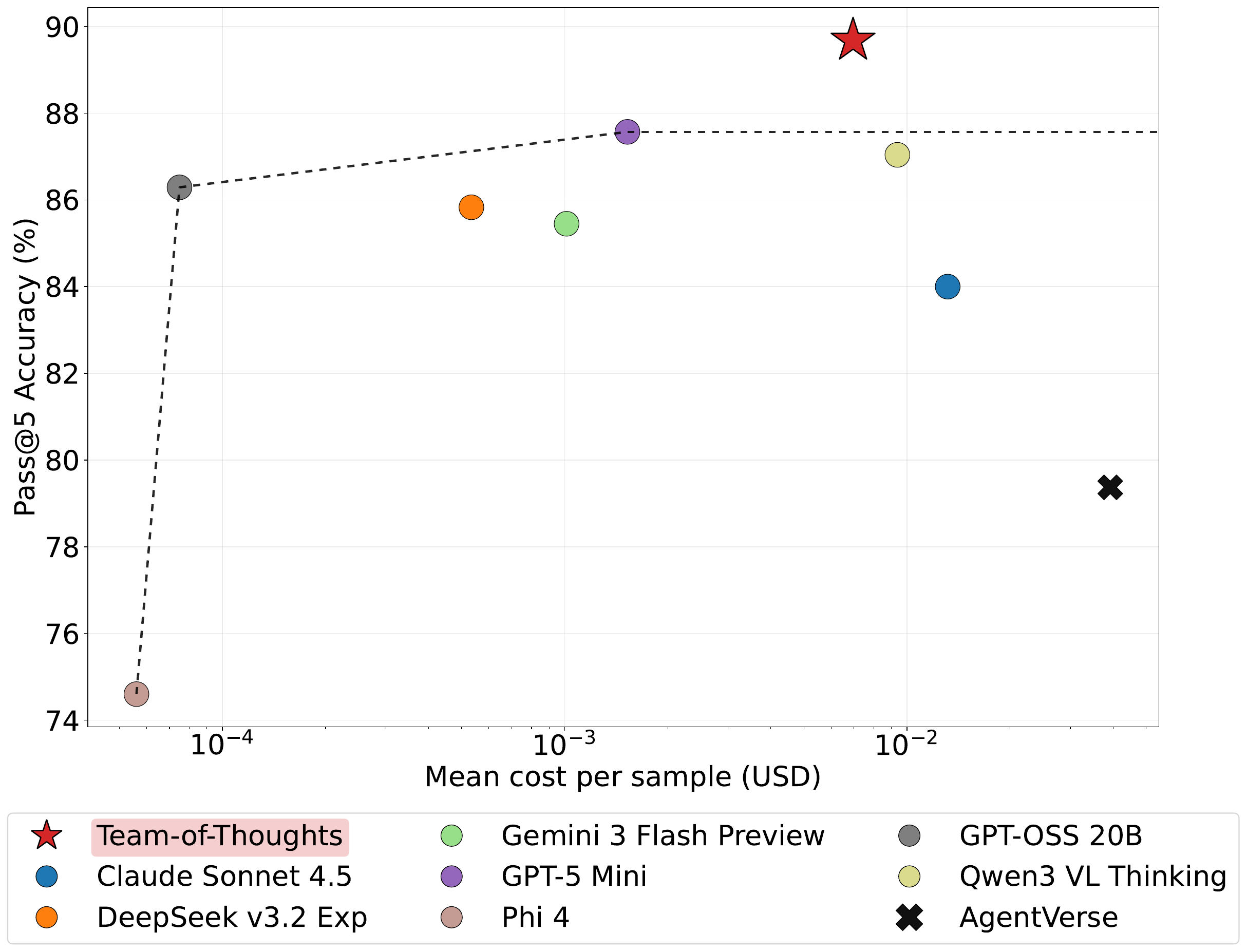}
\caption{Pass@5 accuracy versus cost on MBPP+ for Team-of-Thoughts, single models, and AgentVerse. The dashed line denotes the Pareto front of single models.}
    \label{fig:motivation}
    \vspace{-1em}
\end{figure}

%% file: sections/background.tex
\section{Background}
\label{sec:background}

\subsection{Test-Time Scaling and Reasoning}

Test-time scaling (TTS) enhances performance by allocating additional computational budget during inference. Reasoning frameworks, such as Chain-of-Thought (CoT)~\citep{cot}, Tree-of-Thoughts~\citep{tot}, and Graph-of-Thoughts~\citep{got}, leverage this by decomposing tasks into intermediate steps, enabling structured search over the solution space.

However, a single agent's expressive power is bounded by its fixed parameterization $\theta$. While search-based scaling laws~\citep{ttsscaling, inferencescaling, largelanguagemonkeys} show performance gains, these remain constrained by the model's inherent biases, which may preclude reaching solutions in remote regions of the solution space. As training larger models is often prohibitively expensive, there is a clear need for architectures that scale capability at test-time without re-training.

\subsection{Multi-Agent Systems (MAS)}

Multi-Agent Systems (MAS) address single-model limitations by employing an ensemble $\{\theta_1, \theta_2, \dots, \theta_n\}$. In these systems, agents operate through independent reasoning or collaborative interaction---exchanging intermediate states and debating hypotheses---to synthesize a final output.

Despite their potential, many current MAS frameworks are \textbf{internally homogeneous}, generating ensembles by prompting a single model with varying ``personas''~\citep{agentverse, agentnet, CoA2025}. Because these agents share identical parameterization ($\theta_1 = \theta_2 = \cdots = \theta_n$), they lack the distributional diversity necessary to explore complementary regions of the solution space. Furthermore, these frameworks are often computationally inefficient, requiring multiple rounds of exhaustive reasoning from every agent regardless of their task-specific relevance.

This motivates \textbf{Team-of-Thoughts}, a framework utilizing \textbf{heterogeneous} model priors and a centralized orchestrator to strategically activate specialized agents, optimizing both capability coverage and inference efficiency.

%% file: sections/method.tex
\section{Team of Thoughts: An Efficient Heterogeneous MAS}
\label{sec:method}

\begin{figure*}[t]
    \centering
    \includegraphics[width=\linewidth]{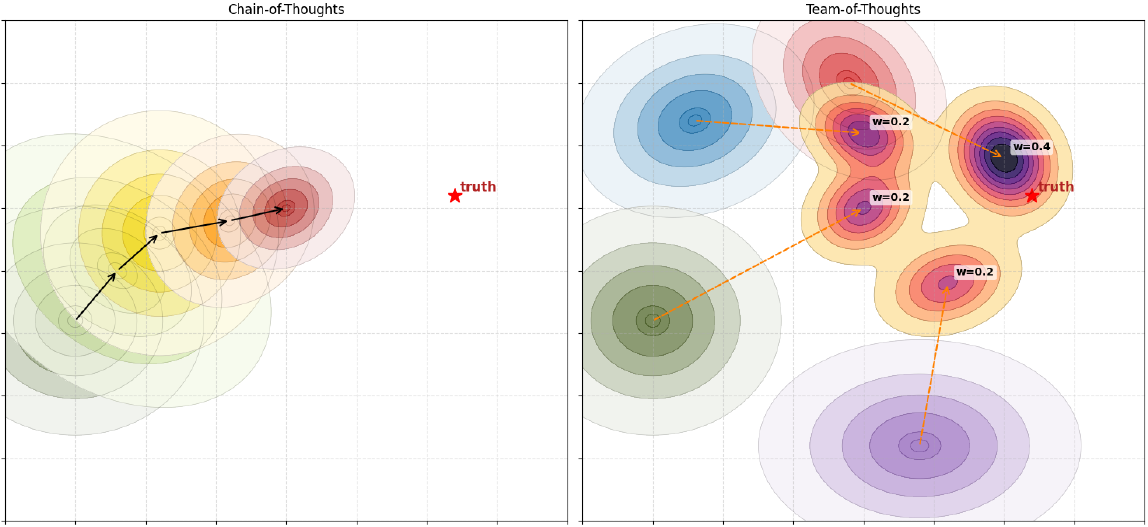}
    \caption{\textbf{Schematic comparison of Chain-of-Thought (CoT) and Team-of-Thoughts.} \textbf{(Left)} Standard agentic reasoning via CoT generates sequential intermediate steps to refine the prediction distribution heuristically. However, on complex tasks, CoT may fail to converge on distant targets. \textbf{(Right)} Team-of-Thoughts utilizes heterogeneous tool agents to explore a broader solution space. During inference, these agents refine their individual distributions through local reasoning, which an orchestrator then prioritizes and aggregates into the final output.}
    \label{fig:GMM}
\end{figure*}

We introduce \textbf{Team-of-Thoughts}, a MAS framework that leverages a suite of heterogeneous agents to maximize capability coverage. The Team-of-Thoughts framework is composed of a central \textbf{orchestrator agent} $p_{\mathrm{orch}}(\cdot \mid D)$ and a diverse ensemble of \textbf{tool agents} $\left\{p_{\theta_i}(\cdot \mid D)\right\}_{i=1}^n$, each characterized by a distinct prediction distribution.

The orchestrator manages the reasoning process by dynamically invoking tool agents based on the requirements of the input query $D$. Specifically, the orchestrator performs three primary functions:
\begin{enumerate}\itemsep0em\parskip0.5em
    \item \textbf{Selection:} Identifying the tool agents best suited to address the specific question of $D$.
    \item \textbf{Evaluation:} Assessing the quality and relevance of the tool agents' outputs.
    \item \textbf{Aggregation:} Synthesizing tool-generated insights with its own reasoning to update the global context.
\end{enumerate}
When invoked, each tool agent $i$ generates a reasoning trajectory $\pmb{\mathrm{Z}}^{(i)} = \{Z_{i,1}, Z_{i,2}, \ldots\}$ and produces a candidate prediction $\hat{X}_i \sim p_{\theta_i}\left(X \mid D, \pmb{\mathrm{Z}}^{(i)}\right)$. The orchestrator then integrates these perspectives to update its context $\pmb{\mathrm{Z}}$ as follows:
\begin{gather}
\label{eq:orch}
Z \sim p_{\text{orch}}\left(\cdot \mid D, \hat{X}_1, \hat{X}_2, \ldots, \hat{X}_k\right) \\
\pmb{\mathrm{Z}} \leftarrow \pmb{\mathrm{Z}} \Vert Z
\end{gather}
where $k \leq n$ denotes the subset of agents active for the given task and $\Vert$ denotes appending the new thinking to the context.

In the rest of this section, we first provide a probabilistic motivation for our approach, and then we detail the implementation of the framework.

\subsection{A Probabilistic View of Team-of-Thoughts}

A task-solving problem comprises an input query $D$ and an unknown target answer $X \in \mathcal{S}$ within the solution space $\mathcal{S}$. We represent the prediction distribution of a language model with parameters $\theta$ as $p_\theta(\cdot \mid D)$. For analytical clarity, we can approximate this distribution as a \textit{multivariate Gaussian}:
$$
p_\theta(\cdot \mid D) \approx \mathcal{N}(\pmb\mu_\theta,\pmb\Sigma_\theta)
$$
where $\pmb\mu$ and $\pmb\Sigma$ denote the mean and covariance matrix, respectively. In inference, we aim to refine this distribution to maximize the likelihood of $X$.

\textbf{Limitations of Single-Agent Reasoning}

Standard agentic frameworks, such as Chain-of-Thought (CoT)~\citep{cot}, attempt to reach the target $X$ by generating a sequence of intermediate reasoning steps $\{Z_1, Z_2, \dots, Z_T\}$. Each step iteratively updates the model's state, shifting the prediction distribution:
$$\begin{gathered}
Z_1 \sim p_\theta(\cdot  \mid  D) \approx  \mathcal{N}\left(\pmb\mu^1_\theta, \pmb\Sigma^1_\theta\right), \\
Z_2 \sim p_\theta(\cdot  \mid  D, Z_1) \approx  \mathcal{N}\left(\pmb\mu^2_\theta, \pmb\Sigma^2_\theta\right), \\
Z_3 \sim p_\theta(\cdot  \mid  D, Z_1, Z_2) \approx  \mathcal{N}\left(\pmb\mu^3_\theta, \pmb\Sigma^3_\theta\right), \\
\vdots  \\
\hat{X} \sim p_\theta(\cdot  \mid  D, Z_1, Z_2, \dots, Z_T) \approx \mathcal{N}\left(\pmb\mu'_\theta, \pmb\Sigma'_\theta\right)
\end{gathered}$$
As illustrated in \Cref{fig:GMM} (left), these steps heuristically ``drift'' the initial distribution toward the target.
However, a single agent's ability to transform the prediction distribution $\mathcal N \left(\pmb\mu'_\theta, \pmb\Sigma'_\theta\right)$ is essentially framed by its initial distribution $\mathcal N\left(\pmb\mu_\theta, \pmb\Sigma_\theta\right)$ defined by the parameterization $\theta$. In complex, high-dimensional tasks where the target $X$ can be significantly distant from the initial mean $\pmb\mu_\theta$, a single reasoning chain may require an impractical number of steps to converge, or may become trapped in local optima within the solution space.

\textbf{Team-of-Thoughts as a Gaussian Mixture}

The Team-of-Thoughts framework overcomes these constraints by employing a Multi-Agent System (MAS) of heterogeneous agents to achieve broader capability coverage. By utilizing an ensemble of $n$ distinct tool agents $E = \left\{p_{\theta_i}\right\}_{i=1}^n$, we effectively initialize the system with multiple \textit{exploratory heads} across the solution space.

As illustrated in \Cref{fig:GMM} (right), the orchestrator agent $p_{\text{orch}}$ selectively invokes these tool agents and integrates their outputs into its global reasoning context. From a probabilistic perspective, the resulting distribution formed by the orchestrator can be modeled as a \textit{Gaussian Mixture Model (GMM)}:
$$
p_{\text{orch}}\left(\cdot \mid D, \mathbf{Z}, \mathbf{\hat{X}}_\text{tools}\right) \approx \sum_{i=1}^n w_i p_{\theta_i}\left(\cdot \mid D, \mathbf{Z}^{(i)}\right)
$$
where $\mathbf{Z}$ represents the orchestrator's reasoning trajectory, $\mathbf{Z}^{(i)}$ is the local reasoning trajectory of the $i$-th tool agent, $\mathbf{\hat{X}}_{\text{tools}}$ is the set of candidate tool predictions, and $w_i$ are mixing weights determined by the orchestrator such that $\sum w_i = 1$.


To implement the weight assignment in a discrete LLM setup, we characterize it as a classification task performed by the orchestrator over the candidate responses:
$$
\mathbf{w} \approx p_\text{orch}\left(\cdot \mid \mathcal C\left(D, \mathbf Z, \mathbf{\hat X}_\text{tools}\right), \mathbf{\hat{X}}_\text{cali}\right)
$$
Here, $\mathcal{C}(\dots)$ denotes a classification query conditioned on the problem and tool generations, while $\mathbf{\hat{X}}_{\text{cali}}$ represents the performance of tool agents on a calibration dataset, detailed in \Cref{sec:self-assess}.

\textbf{Conceptual Nuances}

It is important to note that the GMM serves as a \textit{conceptual framework} rather than a literal mechanical description. The equivalence is approximated because (a) the updated orchestrator distribution is conditioned on the tool agents' \textbf{discrete responses} ($\mathbf{\hat{X}}_{\text{tools}}$) rather than their full continuous distributions ($p_{\theta_i}$), and (b) the weight vector $\mathbf{w}$ is typically \textbf{sparse}, as the orchestrator identifies and selects only a high-confidence subset of $k\le n$ tool agents based on the calibration results.

Despite these approximations, the formulations encapsulate how Team-of-Thoughts dynamically navigates the solution space by prioritizing high-utility distributions. Rather than relying on the ``linear drift'' of single-model reasoning, the orchestrator can effectively ``jump'' to high-probability regions already localized by specialized tool agents. It then refines this synthesized mixture through further reasoning---potentially via iterative tool calls---to produce a definitive, high-confidence answer.

\subsection{Orchestration Calibration}

Orchestration demands a specialized suite of capabilities---including tool comprehension, multi-agent coordination, and trajectory synthesis---that are not uniformly distributed across model architectures. Crucially, a model’s proficiency as a standalone solver does not always correlate with its efficacy as a coordinator. To identify the optimal ``manager'' for our MAS, we introduce an \textbf{Orchestration Calibration} procedure.

We evaluate candidate orchestrators based on their empirical performance in aggregating tool-agent responses within a specific task category $c$, subject to a fixed computational budget. For each candidate $\theta_{\text{cand}}$, we compute a calibration score:
\begin{multline}
\text{Score}(\theta_{\text{cand}}, c) = \\
\frac{\sum_{D \in \mathcal D_{\text{val}}^{(c)}}
\mathbb{I}\left[\hat{X}_{\text{cali}}(D, \theta_\text{cand}) = X_{\text{cali}}(D)\right]}{\left| \mathcal D_{\text{val}}^{(c)}\right|}
\end{multline}
where $\mathcal D_{\text{val}}^{(c)}$ represents the category-specific calibration set, $X_{\text{cali}}(D)$ is the ground-truth answer, and $\hat{X}_{\text{cali}}(D, \theta_{\text{cand}})$ is the prediction generated after the candidate aggregated the available tool agents. The model achieving the highest score is designated as the primary orchestrator for that category.

As demonstrated in \Cref{sec:eval-orch-select}, this process reveals that orchestration capability is often decoupled from model scale or general benchmark rankings. Certain models exhibit superior ``integrative reasoning,'' while others, including some larger models, are more effective as specialized tool agents. These findings validate treating orchestrator selection as a distinct optimization problem rather than defaulting to the largest available model.

\subsection{Agent Specialization via Self-Assessment}
\label{sec:self-assess}

Heterogeneous model families exhibit divergent expertise across task domains due to variations in architecture and post-training procedures. To capitalize on these specialized priors, we introduce a \textbf{Self-Assessment Mechanism} that enables tool agents to characterize their own proficiency profiles. This mechanism consists of three stages:
\begin{enumerate}\itemsep0em\parskip0em
    \item \textbf{Empirical Profiling:} Each tool agent $i$ generates predictions $\hat{X}_i$ and reasoning trajectories $\mathbf{Z}^{(i)}$ on a representative validation set $D_{\text{val}}$:
    $$\hat{X}_i \sim p_{\theta_i}\left(\cdot \mid D, \mathbf{Z}^{(i)}\right), \quad \forall D \in \mathcal D_{\text{val}}$$
    \item \textbf{Performance Quantification:} The system evaluates the accuracy $s_i$ of each agent using an indicator function relative to the ground truth $X(D)$:$$s_i = \frac{1}{\left|D_{\text{val}}\right|} \sum_{D \in \mathcal D_{\text{val}}} \mathbb{I}\left[\hat{X}_i(D) = X(D)\right]$$
    \item \textbf{Qualitative Characterization:} Using its performance data $s_i^{(c)}$ on every the categorized question type $c$ within $\mathcal D_{\text{val}}$, each tool agent generates a ``capability profile.'' This profile is a concise natural language summary detailing the agent's strengths, weaknesses, and reliability for specific task categories.
\end{enumerate}

During inference, these capability profiles are provided to the orchestrator's context. This enables dynamic, priority-based activation: the orchestrator evaluates the incoming query $D$ against the agents' profiles to select only the most compatible tools. This approach facilitates strategic budget allocation---highly proficient agents are prioritized while irrelevant agents are bypassed---contrasting with traditional MAS architectures that invoke all agents regardless of task fit.

%% file: sections/experiment.tex
\section{Experiments}

\input{tables/tab-main-tot-no-cost}

\paragraph{Models and Benchmarks} We evaluate Team-of-Thoughts across seven model families: Claude-Sonnet-4.5~\citep{clause4.5}, GPT-5-mini~\citep{gpt5}, Gemini-3-Flash-Preview~\citep{gemini3}, DeepSeek-V3.2-Exp~\citep{deepseekv3.2}, GPT-OSS-20B~\citep{gptoss}, Qwen3-VL-235B-A22B-Thinking~\citep{qwen3}, and Phi-4~\citep{phi4}. Each tool call activates two tool agents per call. Our evaluation spans mathematical reasoning (AIME2024~\citep{AIME24}, AIME2025~\citep{AIME25}) and code generation (Humaneval+~\citep{humaneval}, MBPP+~\citep{mbpp}, and LiveCodeBench v6~\citep{livecodebench} for Jan--May 2025). 
For calibration, we typically employ a 10\% random sample. However, to account for the limited scale of AIME benchmarks, we adopt a cross-calibration strategy: 50\% of AIME2025 serves as the calibration set for AIME2024, and vice versa. This ensures robust agent profiling while preventing test-set contamination.

\subsection{Performance Analysis}

The Team-of-Thoughts framework integrates tool agents via descriptions derived from their self-assessment profiles. Guided by the calibration results in \Cref{sec:eval-orch-select}, we utilize DeepSeek v3.2 as the orchestrator for mathematical reasoning and GPT-5 Mini for code generation.

As detailed in \Cref{tab:acc-cost-benchmarks}, Team-of-Thoughts demonstrates robust performance across both reasoning and coding domains. On mathematical benchmarks (AIME2024, AIME2025), it achieves superior stability compared to standalone models, consistently attaining higher pass@1 accuracy while remaining competitive at pass@3 and pass@5. In code generation, Team-of-Thoughts achieves state-of-the-art results on MBPP+ across all pass@$k$. On HumanEval+, it trails Claude Sonnet 4.5 marginally at pass@1---a metric increasingly constrained by benchmark saturation---but demonstrates superior scaling by surpassing Claude as the number of allowed generations increases.

For LiveCodeBench, Team-of-Thoughts achieves peak pass@1 performance but marginally lags behind the strongest single-model baseline at pass@3 and pass@5. We hypothesize that default random calibration fails to capture the high task diversity of this benchmark. Validation experiments using a categorized calibration set support this: the strongest models are invoked twice as frequently as a random calibration set, significantly improving overall performance (see Appendix~\ref{apx:calibration_dataset}).

Overall, these results validate the probabilistic intuition behind Team-of-Thoughts. By leveraging heterogeneous agents to cover a broader solution space, the orchestrator ``jumps'' to high-probability regions localized by specialized tools. This strategic prioritization allows the system to synthesize robust answers that transcend the constraints of any single model parameterization.

\subsection{Orchestration Agent Selection}
\label{sec:eval-orch-select}

\input{tables/tab-frontend-selection}

To identify the optimal coordinator, we evaluate candidate models on AIME2024 and MBPP+ under fixed monetary constraints. We normalize budgets by translating costs into model-specific token limits to ensure a rigorous, cost-controlled comparison across providers.
As shown in \Cref{tab:model-benchmark-avg}, orchestration proficiency is task-dependent: DeepSeek v3.2 performs best on the AIME2024 mathematical benchmark, while GPT-5 Mini excels on the MBPP+ coding task. These results indicate that the most effective orchestrator is not necessarily the largest model, but the one with the highest \textbf{orchestrating efficiency} for a specific domain. Consequently, we employ DeepSeek v3.2 for mathematical reasoning and GPT-5 Mini for code generation in our primary analysis.

\subsection{Tool Agent Assessment Strategies}

To optimize agent invocation, we generate granular capability profiles for each tool agent. We evaluate three distinct selection policies to determine how these profiles are best constructed and utilized:

\textbf{Random Invocation (Baseline):} Tool models are sampled uniformly from the ensemble without prior profiling and domain-specific weighting.

\textbf{Orchestrator-Led Assessment:} A centralized model (DeepSeek-v3.2 for math; GPT-5 Mini for coding) evaluates the tool agents on a calibration set to map their competencies and common failure modes.

\textbf{Agent Self-Assessment:} Each tool agent performs a self-audit. Provided with the task, its own reasoning trajectories, and the ground truth, the agent identifies the specific skills required and critiques its own performance.

These profiles allow the orchestrator to dynamically adjust invocation preference based on task-agent compatibility. As detailed in \Cref{tab:strategy-comparison}, all structured selection policies significantly outperform the orchestrator-only baseline (``Single''). Among these, the \textbf{self-assessment} policy achieves the highest overall accuracy. 
A key advantage of self-assessment is that it decouples capability profiling from the orchestrator's internal biases, providing a more objective and stable representation of agent expertise. Consequently, we adopt self-assessment as the default selection strategy.

\input{tables/tab-backend-selection}

\subsection{Tool Response Aggregation Strategies}

\input{tables/tab-include-tool-traces}

In the Team-of-Thoughts framework, an activated tool agent $\theta_i$ is prompted with a query $D_{\text{call}}$. The agent generates a sequence of intermediate reasoning steps $\mathbf{Z}^{(i)} = \left[Z_1^{(i)}, \dots, Z_T^{(i)}\right]$ where $Z_t^{(i)} \sim p_{\theta_i}(\cdot \mid D_{\text{call}}, \mathbf{Z}_{1:t-1}^{(i)})$, reaching a final output $\hat{X}_i$. We evaluate two distinct protocols for information exchange between the tool agent and the orchestrator: (1) returning only the final solution $\hat{X}_i$, or (2) providing the complete reasoning trajectory $\mathbf{Z}^{(i)}$ alongside the answer.

As summarized in \Cref{tab:include-traces}, our results indicate that incorporating full reasoning trajectories consistently degrades orchestration performance across all benchmarks. This performance drop suggests that verbose trajectories introduce significant contextual noise, which can distract or mislead the orchestrator through intermediate logical errors or irrelevant sub-steps.
These findings demonstrate that for complex multi-agent synthesis, minimal tool responses enhance the orchestrator's ability to integrate heterogeneous information. By filtering out potentially fallible intermediate steps, the orchestrator maintains a cleaner, more reliable global context, leading to more robust decision-making and higher task accuracy.

\subsection{Scaling Dynamics of Tool-Agent Ensemble}
\label{sec:Scaling Dynamics of Tool-Agent Ensemble}
While Team-of-Thoughts exhibits high efficacy, the performance of Multi-Agent Systems (MAS) does not scale monotonically with ensemble size. We investigate the scaling behavior of Team-of-Thoughts relative to two variables: the size of the available ensemble ($|E|$) and the number of activated agents per call ($k$). To test the limits of this architecture, we expanded the ensemble $E$ to include the 100 most popular models on OpenRouter and evaluated performance on MBPP+.

\input{tables/tab-100-models-cmp-act-avail-25}

\textbf{Impacts of Activation Density:} We first examine whether the orchestrator benefits from a higher volume of concurrent agent responses. Fixing the available ensemble at $|E|=25$, we varied the number of activated models $k$ from 2 to 16. As shown in \Cref{tab:100-models-cmp-act-avail-25}, increasing $k$ yields negligible gains and eventually leads to performance degradation. Notably, Pass@1 accuracy drops significantly once $k$ reaches 8.
The narrowing performance gap at Pass@3 and Pass@5 suggests that while higher activation density $k/n$ \textbf{increases run-time variance}---widening the distribution to cover the correct answer across multiple attempts---it simultaneously \textbf{compromises the robustness} of the single-shot (Pass@1) prediction.

\input{tables/tab-100-models-cmp-avail-act-2and8}

{Ensemble Breadth and Selection Overload.} We further analyze whether the orchestrator's selection improves when provided with a broader pool of candidates. Sweeping over ensemble sizes for $k=2$ and $k=8$, \Cref{tab:100-models-cmp-avail-act-2and8} reveals that while $k=2$ remains stable across ensemble sizes, $k=8$ exhibits a clear performance decay as $|E|$ increases. To isolate the root cause, we define the \textbf{Selection Agreement} between these two configurations:
\[
\text{Agreement} = \frac{1}{\left|D\right|} \sum_{i=1}^{\left|D\right|} \mathbb{I}\left[S_2^{(i)} \subseteq S_8^{(i)}\right]
\]
where $S_k^{(i)}$ denotes the set of tool agents invoked for the $i$-th question at activation level $k$. As shown in the final column of \Cref{tab:100-models-cmp-avail-act-2and8}, Agreement sharply declines once $|E| > 50$. This suggests a selection logic overload phenomenon: the orchestrator's ability to identify optimal tools is overwhelmed by excessive options, leading to invocations of suboptimal agents.

Even though Agreement remains high when $|E| \leq 25$, a Pass@1 performance gap persists. This indicates that even with ``optimal'' tool selection, the high density of responses (eight vs.\ two) introduces excessive variance into the aggregation process, exceeding the orchestrator's synthesis capacity.

Our findings demonstrate that MAS performance is constrained by an orchestration bottleneck. Scaling both activation density and ensemble breadth eventually degrades performance by overwhelming the orchestrator's selection and aggregation logic. Conversely, with a calibrated activation count (e.g.\ $k=2$), the orchestrator can reliably extract high-utility responses from an arbitrarily large ensemble, producing stable, high-confidence solutions.

%% file: tables/tab-main-tot-no-cost.tex
\definecolor{lightgreen}{RGB}{200, 255, 200}
\definecolor{lightred}{RGB}{225, 200, 200}

\begin{table*}[t]
\centering
\caption{
\textbf{Performance comparison of Team-of-Thoughts across five benchmark tasks.} Success rates (\%) are reported using pass@1, 3, and 5. Team-of-Thoughts is evaluated against individual model baselines and state-of-the-art multi-agent methods. The \textit{Theoretical Limit} represents the oracle upper bound achieved by selecting the best result among all individual models for each problem.
}
\label{tab:acc-cost-benchmarks}
\resizebox{\linewidth}{!}{%

\centering
\small
\begin{tabular}{lccccccccccccccc}
\toprule
 & \multicolumn{3}{c}{AIME 2024} 
 & \multicolumn{3}{c}{AIME 2025} 
 & \multicolumn{3}{c}{Humaneval+} 
 & \multicolumn{3}{c}{MBPP+} 
 & \multicolumn{3}{c}{LiveCodeBench v6} \\
\cmidrule(lr){2-4} \cmidrule(lr){5-7} \cmidrule(lr){8-10} \cmidrule(lr){11-13} \cmidrule(lr){14-16}
Model 
& pass@1 & @3 & @5 
& @1 & @3 & @5
& @1 & @3 & @5
& @1 & @3 & @5
& @1 & @3 & @5 \\
\midrule

Claude Sonnet 4.5
& 84.00 & 86.67 & 86.67
& 74.67 & 87.67 & 90.00
& \textbf{95.61} & 96.34 & 96.34
& 81.80 & 84.29 & 85.45
& 48.24 & 55.33 & 57.69 \\

GPT-5-mini
& 88.67 & 93.33 & 93.33
& 73.33 & 83.50 & 86.11
& 92.68 & 94.88 & 95.12
& 82.38 & 86.19 & 87.57
& 64.84 & 74.40 & 77.20 \\

Gemini 3 Flash
& 86.67 & 90.67 & 93.33
& 71.33 & 81.33 & 86.67
& 96.46 & 96.95 & 96.95
& 84.44 & 85.13 & 85.45
& 72.53 & 75.71 & 77.47 \\

Deepseek v3.2 Exp
& 94.67 & \textbf{96.67} & \textbf{96.67}
& 94.00 & 98.67 & \textbf{100.00}
& 91.26 & 95.88 & 96.65
& 80.20 & 84.59 & 85.83
& 76.48 & \textbf{86.65} & \textbf{88.46} \\

GPT OSS 20b
& 75.33 & 88.00 & 90.00
& 73.33 & 83.50 & 86.11
& 89.23 & 95.12 & 95.63
& 79.23 & 84.88 & 86.29
& 46.04 & 54.34 & 58.24 \\

Qwen3-vl 235B
& 93.33 & 93.33 & 93.33
& 89.44 & 91.67 & 92.78
& 93.19 & 94.70 & 95.02
& 83.20 & 86.24 & 87.04
& 67.47 & 75.77 & 78.57 \\

Phi-4
& 13.33 & 18.00 & 20.00
& 10.67 & 15.67 & 16.67
& 74.80 & 80.55 & 82.83
& 63.07 & 72.51 & 74.60
& 24.84 & 27.69 & 28.02 \\

\midrule

\textit{Theoretical Limit}
& 96.67 & -- & --
& 100.00 & -- & --
& 100.00 & -- & --
& 100.00 & -- & --
& 90.50 & -- & -- \\

Majority Voting
& 93.33 & -- & --
& 93.33 & -- & --
& N/A & N/A & N/A
& N/A & N/A & N/A
& N/A & N/A & N/A \\

AgentVerse
& 80.00 & -- & --
& 60.00 & -- & --
& 91.46 & -- & --
& 79.37 & -- & --
& 65.93 & -- & -- \\



\midrule
Team-of-Thoughts
& \textbf{96.00} & \textbf{96.67} & \textbf{96.67}
& \textbf{95.33} & \textbf{99.33} & \textbf{100.00}
& 95.33 & \textbf{97.47} & \textbf{98.07}
& \textbf{85.93} & \textbf{88.84} & \textbf{89.68}
& \textbf{77.91} & 85.05 & 87.91 \\

\bottomrule
\end{tabular}
}
\end{table*}

%% file: tables/tab-frontend-selection.tex
\begin{table}
\centering
\caption{\textbf{Orchestration performance across candidate models and budget constraints.} Calibration accuracy (\%) is evaluated on AIME2024 and MBPP+ benchmarks under two distinct budgetary tiers (USD). Avg denotes the mean accuracy across all evaluated settings. 
}
\label{tab:model-benchmark-avg}
\resizebox{\columnwidth}{!}{%
\begin{tabular}{lcccccc}
\toprule
\textbf{Model} 
& \multicolumn{3}{c}{\textbf{AIME2024}} 
& \multicolumn{3}{c}{\textbf{MBPP+}} \\
\cmidrule(lr){2-4} \cmidrule(lr){5-7}
Budget Cost (\$) & 0.03 & 0.02 & Avg & 0.03 & 0.02 & Avg \\
\midrule
Claude Sonnet 4.5
& 86.67 & 46.67 & 66.67
& 78.38 & 78.38	& 78.38 \\

GPT-5 Mini
& 86.67 & 93.33 & 90.00
& 86.49 & 83.78	& \textbf{85.14} \\

Gemini 3 Flash
& 86.67 & 40.00 & 63.34
& 83.78 & 83.78	& 83.78 \\

DeepSeek v3.2
& 93.33 & 93.33 & \textbf{93.33}
& 81.08 & 81.08	& 81.08 \\

GPT-OSS 20B
& 80.00 & 80.00 & 80.00
& 75.68 & 75.68 & 75.68 \\

Qwen3-vl 235B
& 33.33 & 20.00 & 26.67
& 78.38 & 78.38 & 78.38 \\

Phi-4
& 33.33 & 33.33 & 33.33
& 72.97 & 72.97 & 72.97 \\
\bottomrule
\end{tabular}
}
\vspace{-1em}
\end{table}

%% file: tables/tab-backend-selection.tex
\begin{table}[t]
\centering
\caption{\textbf{Performance evaluation of tool agent assessment strategies.} Comparison of Pass@1 accuracy (\%) with no tool agent (``Single''), Orchestrator-Led Assessment, Agent Self-Assessment, and Random Invocation across AIME2024 and MBPP+ benchmarks.
}
\label{tab:strategy-comparison}
\resizebox{\linewidth}{!}{%
\begin{tabular}{lcccc}
\toprule
\textbf{Benchmark} & \textbf{Single} & \textbf{Self-Assess} & \textbf{Orch-Based} & \textbf{Random} \\
\midrule


AIME2024 
& 94.67
& \textbf{96.00}
& 92.67
& 94.00 \\

MBPP+ 
& 82.38
& \textbf{85.93}
& 85.77
& 84.44 \\

\bottomrule

\end{tabular}
}
\end{table}

%% file: tables/tab-include-tool-traces.tex
\begin{table}[!t]
\centering
\caption{\textbf{Performance of orchestrator aggregation strategies.} Comparison of task accuracy (\%) achieved when incorporating versus excluding the intermediate reasoning traces of tool agents.}
\label{tab:include-traces}
\resizebox{\linewidth}{!}{%
\begin{tabular}{lcccccc}
\toprule
               & \multicolumn{3}{c}{AIME 2025} & \multicolumn{3}{c}{MBPP+} \\
\cmidrule(lr){2-4} \cmidrule(lr){5-7}
Config         & pass@1   & @3   & @5  & pass@1  & @3 & @5 \\
\midrule
No traces      & \textbf{95.33} & \textbf{99.33} & \textbf{100.00}  & \textbf{85.93}   & \textbf{88.84}  & \textbf{89.68}  \\
Include traces & 78.67    & 93.33    & 96.67   & 80.90   & 85.61  & 87.04  \\
\bottomrule
\end{tabular}%
}
\vspace{-1em}
\end{table}

%% file: tables/tab-100-models-cmp-act-avail-25.tex
\begin{table}[h]
\centering
\caption{
Performance under different numbers of activated tool agents, with 25 total tool agents available.
}
\label{tab:100-models-cmp-act-avail-25}
\resizebox{\linewidth}{!}{%
\begin{tabular}{lccc}
\toprule
Activated tools & Pass@1  & Pass@3  & Pass@5  \\
\midrule
2            & 82.75\% & 85.58\% & 86.77\% \\
4            & 82.96\% & 85.32\% & 85.98\% \\
8            & 60.32\% & 83.76\% & 86.24\% \\
16           & 60.63\% & 83.10\% & 85.71\% \\
\bottomrule
\end{tabular}%
}
\end{table}

%% file: tables/tab-100-models-cmp-avail-act-2and8.tex
\begin{table}[ht]
\centering
\caption{Orchestrator performance and tool selection consistency (Agreement) across varying ensemble sizes. Performance is compared between configurations activating two and eight agents per tool call.}
\label{tab:100-models-cmp-avail-act-2and8}
\resizebox{\linewidth}{!}{%
\begin{tabular}{lccccccc}
\toprule
& \multicolumn{3}{c}{\textbf{Activate 2}} & \multicolumn{3}{c}{\textbf{Activate 8}} & \textbf{Agreement} \\
\cmidrule(lr){2-4} \cmidrule(lr){5-7} \cmidrule(lr){8-8}
Avail Tools & Pass@1 & @3 & @5 & Pass@1 & @3 & @5 & $\mathcal{A}$ (\%) \\
\midrule
10  & 82.70 & 86.32 & 87.57 & 74.97 & 85.26 & 85.98 & 90.74 \\
25  & 82.75 & 85.58 & 86.77 & 60.32 & 83.76 & 86.24 & 91.80 \\
50  & 82.43 & 86.59 & 87.83 & 54.76 & 82.17 & 86.51 & 61.64 \\
75  & 82.49 & 87.09 & 88.36 & 63.65 & 84.21 & 86.24 & 75.13 \\
100 & 82.49 & 86.56 & 87.57 & 61.43 & 84.13 & 86.51 & 71.43 \\
\bottomrule
\end{tabular}%
}
\vspace{-1em}
\end{table}

%% file: sections/related.tex
\section{Related Work}

\textbf{Test-Time Scaling (TTS).}
TTS establishes that increasing inference-time computation can yield performance gains comparable to scaling model parameters~\citep{ttsscaling, inferencescaling}. Approaches generally follow two trajectories: (1) \textbf{parallel sampling}, such as Best-of-N~\citep{largelanguagemonkeys}, which uses verification to select optimal outputs from a broad search space; and (2) \textbf{sequential reasoning}, which deepens the reasoning topology. This includes linear Chain-of-Thought (CoT)~\citep{cot} and non-linear frameworks like Tree of Thoughts (ToT)~\citep{tot} and Graph of Thoughts (GoT)~\citep{got}. These methodologies enable complex operations like backtracking and information aggregation, mimicking human ``System 2'' cognition.

\textbf{Multi-Agent Systems (MAS).} 
Inspired by human dynamics, MAS utilizes agent ensembles to enhance task reliability and role-playing capabilities~\citep{Simulacra, camel}. Early frameworks leveraged conversational flows~\citep{autogen} or Standard Operating Procedures (SOPs) to reduce hallucinations in software engineering~\citep{chatdev, metagpt}. More recent research emphasizes dynamic workflows; for instance, AgentVerse~\citep{agentverse} introduces ``expert recruitment,'' while AgentNet~\citep{agentnet} employs autonomous graph topologies for task decomposition. Theoretically, the ``Society of Thoughts''~\citep{SoT} suggests that even single-model reasoning benefits from simulating diverse internal perspectives.

In contrast to these frameworks, which often rely on \textbf{homogeneous} model ensembles, our \textbf{Team-of-Thoughts} framework explicitly exploits the divergence inherent in \textbf{heterogeneous} model priors. By dynamically selecting specialized agents and aggregating their distinct reasoning paths, we facilitate more robust decision-making than systems restricted to a single model family.

%% file: sections/conclusion.tex
\section{Conclusion}
We propose the Team-of-Thoughts framework, which explicitly leverages the skill diversity of a group of heterogeneous agent models. By dynamically selecting the most suitable orchestrator and making skill-dependent tool calls that can be executed in parallel, Team-of-Thoughts overcomes the limitations of fixed-role, sequential multi-agent systems, achieving superior performance across reasoning and code generation tasks.
Our analysis and empirical experiment across reasoning and code generation tasks demonstrates that this approach consistently achieves more efficient cost usage and higher accuracy compared to both single-model baselines and multi-agent systems with homogeneous backbone models. This work highlights a novel paradigm for enabling heterogeneous models to collaborate, paving the way for future exploration of multi-round, complex multi-agent tasks.

\section*{Limitations}
The proposed Team-of-Thoughts framework relies on a central LLM as an orchestrator, which introduces a performance bottleneck as the agent pool scales. In \Cref{sec:Scaling Dynamics of Tool-Agent Ensemble}, we characterize the limitations of this centralized architecture when scaled to 100 tool-agents. Specifically, we observe two primary failure modes: The orchestrator often fails to navigate expansive tool selection spaces, leading to suboptimal tool-calling sequences. Large-scale tool orchestration induces high variance in output quality as show in pass@1 performance, suggesting a lack of robustness in the model's iterative reasoning process under high context load. These results suggest that the scalability of the system is currently constrained by the state of frontier models. Future advancements in long-context reasoning and hierarchical information aggregation will be essential to fully realize the potential of massive tool-agent ensembles.

%% file: sections/appendix.tex
\section{Experiment Setup Details}

\subsection{Additional Experiment Setup}
\label{apx:exp:setup}

\paragraph{Main experiment setting} We evaluated Team-of-Thoughts MAS across a diverse suite of seven model families, comprising three closed-source models: Claude-Sonnet-4.5~\citep{clause4.5}, GPT-5-mini~\citep{gpt5}, and Gemini-3-Flash-Preview~\citep{gemini3} and four open-source models: DeepSeek-V3.2-Exp~\citep{deepseekv3.2}, GPT-OSS-20B~\citep{gptoss}, Qwen3-VL-235B-A22B-Thinking~\citep{qwen3}, and Phi-4~\citep{phi4}. Two tool agents are activated on each tool call. 

Our assessment spanned two domains: mathematical reasoning (AIME2024~\citep{AIME24}, AIME2025~\citep{AIME25}) and code generation (Humaneval+~\citep{humaneval}, MBPP+~\citep{mbpp}, and LiveCodeBench v6~\citep{livecodebench} with problems released between 2025/01/01 and 2025/05/01). Unless stated otherwise, we set a standardized context window for each tool-agent: 20,000 tokens for AIME tasks and 4,096 tokens for coding tasks. For the orchestrator, we used a 16,384 token context window across all tasks to ensure sufficient capacity for processing tool descriptions and making informed selection and reasoning decisions. For reasoning models, we applied the default ``medium'' effort setting, capping the reasoning token budget at 50\% of the maximum generation length to ensure consistent comparisons across all baselines.

By default, we randomly sample 10\% of the target tasks to construct a calibration dataset. However, for AIME2024 and AIME2025, 10\% corresponds to only three data points, which is too small to provide a reliable characteristic modeling. We instead sample 50\% of the problems from AIME2025 and use them as the calibration dataset for evaluating AIME2024 and vice versa for AIME2025. This cross-calibration setup ensures sufficient calibration data while avoiding exposure of the test data.

\paragraph{Orchestration selection} In the orchestration agent selection experiment, we judge the performance of the orchestrator by activating all tool agents. The max generation token is set based on each agent's cost, ensuring it will not exceed the cost budgets.

\paragraph{Scaling tool agents experiment setting} 
For this analysis, we select the top 100 OpenRouter models ranked by popularity that provide a context length greater than 16K tokens and cost less than \$1 USD per million input tokens. The orchestrator context window is set as 16,384 unless stated otherwise. Due to varying max context lengths of different models, we employ Orchestrator-Led Assessment using GPT-5 Mini. The orchestrator is provided with the tool agents with top single-model performance on MBPP+ when a subset of the models is made available.

\subsection{AgentVerse Setup} 

We employ GPT-5-Mini~\citep{gpt5} as the backbone language model of agents in AgentVerse~\citep{agentverse}. We use a maximum token limit of 512 for the role assigner agent, and 4,096 for the rest of the agents. For invalid agent outputs, such as invalid role assignments, unparseable answers, or errors in code, AgentVerse retries generation a limited number of times: 10 times on math tasks and 1,000 times on coding tasks.


\section{Impact of Calibration Dataset Choice}
\label{apx:calibration_dataset}
We present additional results using a categorized calibration set. To construct the categorized calibration dataset, we first randomly sampled 10\% of the LiveCodeBench v6 dataset as before, resulting in 18 questions. We then used GPT-5 to group these questions according to the primary algorithmic concept involved, such as Arrays / Data Structures, Strings / Sequences, Math / Logic, Grid / Matrix, Simulation / Greedy, Knapsack / Optimization, and Intervals / Range. From each category, one representative question was selected to form the final calibration dataset.

\begin{table}
\centering
\caption{Performance in pass@k on LiveCodeBench v6 using different calibration construction strategies.}
\label{tab:livecodebench_results}
\resizebox{\linewidth}{!}{%
\begin{tabular}{lccc}
\toprule
\textbf{Construction Method} & \textbf{pass@1} & \textbf{pass@3} & \textbf{pass@5} \\
\midrule
Random Sample      & 77.91\% & 85.05\% & 87.91\% \\
Categorized Sample & 81.65\% & 85.93\% & 86.81\% \\
\bottomrule
\end{tabular}
}
\end{table}

\input{figures/fig_apx_categorised_vs_random}

As shown in \Cref{tab:livecodebench_results}, the categorized calibration dataset leads to a noticeable improvement in pass@1 accuracy. \Cref{fig:apx:calibration} compares the distribution of Team-of-Thoughts tool selections under the two calibration strategies. When using the categorized calibration set, the system selects the strongest single model approximately twice as often as when using the random calibration set. This explains the improved pass@1 performance.

However, the categorized calibration set also results in lower diversity in tool selection across generations. Consequently, the pass@5 performance is slightly lower than that obtained with the random calibration dataset. This suggests a trade-off between bias and variance: while the categorized calibration set biases the system toward stronger single-model decisions, it reduces exploration across different tools.

Designing calibration datasets that better balance this trade-off remains an interesting direction for future work.

\section{Longer Context Window with 100-Tool Agent Ensemble}
\label{apx:Longer context window on 100-Agent Team}

\input{tables/tab-100-models-cmp-maxtoken-act-8-avail-25}

To verify whether the performance decline with eight activated agents stems from context window limitations, we evaluated the 25-model ensemble using extended orchestration contexts. As shown in \Cref{tab:100-models-cmp-max-token}, increasing context length yielded no performance gains. This confirms that the bottleneck resides in the orchestrator's aggregation capability rather than spatial context constraints.

\section{Experiment Prompt}
\label{prompt:assessment_prompt}
Here is an example of the language model-based tool agent assessment prompts.

\begin{tcolorbox}[colback=blue!5!white, colframe=blue!75!black, title=Assessment Prompt, breakable]
\textbf{Instructions:} You will be provided with a series of problems. For each problem, the Subject Agent has provided an answer. Some are correct, and some are incorrect.

\textbf{Part 1: Per-Problem Analysis} \\
For every problem provided, generate a structured audit containing:

\begin{enumerate}
    \item \textbf{Taxonomy:} Classify the problem type (e.g., Arithmetic, Logical Reasoning, Creative Writing, Coding) and specific skill required.
    \item \textbf{Performance Verdict:} Clearly state [PASS] or [FAIL].
    \item \textbf{Gap Analysis:}
    \begin{itemize}
        \item \textit{If Correct:} Briefly explain why the agent succeeded (e.g., ``Good step-by-step reasoning,'' ``Robust knowledge retrieval'').
        \item \textit{If Incorrect:} Pinpoint exactly \textit{where} and \textit{why} the agent failed. Was it a calculation error? A logic jump? A hallucination? A misunderstanding of constraints? Compare the Subject's logic to the Ground Truth.
    \end{itemize}
\end{enumerate}

\textbf{Part 2: Executive Summary} \\
After analyzing all problems, synthesize a ``Model Persona'' profile:

\begin{enumerate}
    \item \textbf{Core Competencies:} List specific categories where the agent consistently succeeds.
    \item \textbf{Blind Spots \& Failure Modes:} Describe the patterns in the agent's errors (e.g., ``The agent struggles with negative integers,'' or ``The agent is verbose but inaccurate'').
    \item \textbf{Final Verdict:} A 2-sentence summary of the agent's reliability.
\end{enumerate}

---

\textbf{Input Data:}

\begin{itemize}
    \item Problem 1: \textit{Question, Subject Agent Answer, Ground Truth/Solution...}
    \item Problem 2: \textit{...repeat for all samples...}
\end{itemize}

---

\textbf{COMMAND:} \\
Based on the data above, proceed with the \textbf{Per-Problem Analysis} followed by the \textbf{Executive Summary}.

\textbf{Reminders:}
\begin{enumerate}
    \item \textbf{Be Specific:} Do not just say ``The agent failed.'' Identify \textit{why} (e.g., Logic Error vs. Calculation Error).
    \item \textbf{Be Critical:} Compare the Subject Answer against the Ground Truth rigorously.
    \item \textbf{Format:} Use the headers \texttt{\#\# Part 1: Per-Problem Analysis} and \texttt{\#\# Part 2: Executive Summary}.
\end{enumerate}

\textbf{GENERATE REPORT:}

\end{tcolorbox}

%% file: figures/fig_apx_categorised_vs_random.tex
\begin{figure}
    \centering
    \includegraphics[width=\linewidth]{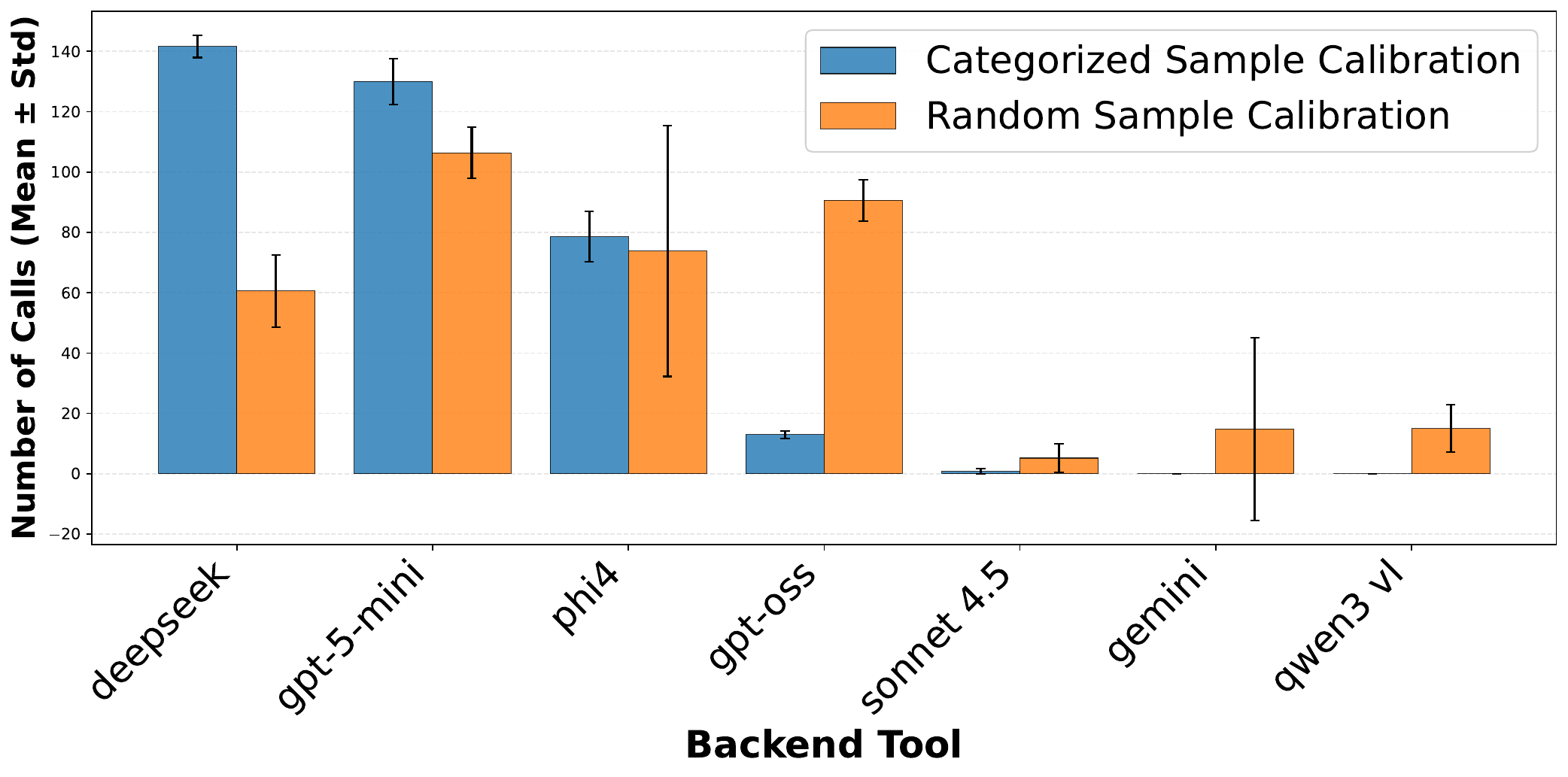}
    \caption{Comparing the times of tool agent invocations between two calibration dataset construction methods.}
    \label{fig:apx:calibration}
\end{figure}

%% file: tables/tab-100-models-cmp-maxtoken-act-8-avail-25.tex
\begin{table}[ht]
\centering
\caption{
Performance under different orchestrator max context lengths, with eight of 25 available tool agents activated on each tool call.
}
\label{tab:100-models-cmp-max-token}
\resizebox{\linewidth}{!}{%
\begin{tabular}{lccc}
\toprule
Orch context len & Pass@1  & Pass@3  & Pass@5  \\
\midrule
16K         & 60.32\% & 83.76\% & 86.24\% \\
32K         & 59.79\% & 82.99\% & 85.71\% \\
64K         & 59.42\% & 83.39\% & 86.51\% \\
128K        & 60.32\% & 82.96\% & 85.45\% \\
\bottomrule
\end{tabular}%
}
\end{table}